\documentclass[conference]{IEEEtran}
\IEEEoverridecommandlockouts
\usepackage{cite}
\usepackage{amsmath,amssymb,amsfonts}
\usepackage{algorithmic}
\usepackage{graphicx}
\usepackage{textcomp}
\usepackage{xcolor}
\usepackage{hyperref}
\usepackage{subfigure}
\usepackage{multirow}

\def\BibTeX{{\rm B\kern-.05em{\sc i\kern-.025em b}\kern-.08em
    T\kern-.1667em\lower.7ex\hbox{E}\kern-.125emX}}

\newcommand{\eg}{\textit{e.g.}}
\newcommand{\ie}{\textit{i.e.}}

\begin{document}

\title{Imbalanced Graph Classification with Multi-scale Oversampling Graph Neural Networks}
\author{
    \IEEEauthorblockN{Rongrong Ma\textsuperscript{\rm 1},
    Guansong Pang\textsuperscript{\rm 2}\thanks{Guansong Pang is the corresponding author.},
    Ling Chen\textsuperscript{\rm 1}}
    \IEEEauthorblockA{\textsuperscript{\rm 1}Faculty of Engineering and Information Technology, University of Technology Sydney, Sydney, Australia\\
    \textsuperscript{\rm 2}School of Computing and Information Systems, Singapore Management University, Singapore\\
    Rongrong.ma-1@student.uts.edu.au, gspang@smu.edu.sg, ling.chen@uts.edu.au} \\
}

\maketitle

\begin{abstract}
One main challenge in imbalanced graph classification is to learn expressive representations of the graphs in under-represented (minority) classes. Existing generic imbalanced learning methods, such as oversampling and imbalanced learning loss functions, can be adopted for enabling graph representation learning models to cope with this challenge. However, these methods often directly operate on the graph representations, ignoring rich discriminative information within the graphs and their interactions. To tackle this issue, we introduce a novel multi-scale oversampling graph neural network (MOSGNN) that learns expressive minority graph representations based on intra- and inter-graph semantics resulting from oversampled graphs at multiple scales - subgraph, graph, and pairwise graphs. It achieves this by jointly optimizing subgraph-level, graph-level, and pairwise-graph learning tasks to learn the discriminative information embedded within and between the minority graphs. Extensive experiments on 16 imbalanced graph datasets show that MOSGNN i) significantly outperforms five state-of-the-art models, and ii) offers a generic framework, in which different advanced imbalanced learning loss functions can be easily plugged in and obtain significantly improved classification performance. Code is available at: \url{https://github.com/RongrongMa/MOSGNN}.
\end{abstract}

\begin{IEEEkeywords}
graph classification, imbalanced learning, oversampling, graph neural networks
\end{IEEEkeywords}

\section{Introduction}
Imbalanced graph classification, which aims to classify graphs with largely skewed or imbalanced class distribution, is an important research topic due to its broad applications in, \eg, recognizing frauds or spams in social networks~\cite{liu2021pick}, locating web application bugs~\cite{aggarwal2010graph},
identifying properties of compounds with different functions~\cite{gupta2021graph}, 
and predicting activities of chemical compounds in bioassay test~\cite{ pan2013graph}.  

One main challenge in imbalanced graph classification is to learn expressive representations capturing discriminative local (\ie, node/subgraph levels) and global (\ie, graph level) knowledge of the graphs in under-represented (minority) classes. In recent years, graph neural networks (GNNs) have demonstrated substantially improved performance on graph representation learning over traditional kernel-based methods \cite{kipf2016semi,xu2018powerful}.
However, they are not designed to tackle the class imbalance problem in graph classification. 

There are numerous conventional imbalanced learning methods, such as oversampling and various loss functions (a.k.a. sample weighting methods) \cite{he2009learning,lin2017focal,tan2020equalization,cao2019learning,menon2020long,kini2021label,zhang2021deep}. Most of them are data-agnostic approaches, so they can often be combined with GNNs directly, \eg, oversampling in training batches or weighting the training samples to GNNs, for class-imbalanced graph representation learning. 
The resulting graph-level representations, however, ignore diverse discriminative information within the graphs and their interactions, especially for the minority graphs. As a result, they are often lack of discriminative power for the classification task.

Recently there have been a number of deep methods designed specifically for imbalanced classification in the graph domain, \eg, DR-GCN~\cite{shi2020multi}, RECT~\cite{wang2020network}, GraphSMOTE~\cite{zhao2021graphsmote}, ImGAGN~\cite{qu2021imgagn} and GraphENS~\cite{park2021graphens}, but they concentrate on imbalanced node classification. 
Limited studies are focused on graph-level imbalanced classification. Particularly, a few data augmentation-based methods, such as random edge/node elimination and mixup \cite{wang2021mixup,han2022g}, have been explored to capture intra-/inter-graph interactions and generate more graphs to enrich the GNN-based representations of minority graphs. However, it is difficult to perform meaningful intra-graph augmentation or interpolations between graphs due to their non-Euclidean nature and the lack of domain knowledge. Wang et al. \cite{wang2022imbalanced} construct a graph of graphs (GoG) on the oversampled and augmented graphs and perform GoG propagation to aggregate neighboring graph representations for imbalanced graph classification. Nevertheless, as the GoG is constructed on kernel similarity, its time complexity increases drastically with dataset size (\ie, polynomial w.r.t. the number of graphs and cubic w.r.t. the maximum number of nodes in a graph), and thus, it is prohibitively computationally costly for large datasets. 

\begin{figure}[t]    
    \centering
    \subfigure[]{\label{motivation-pairwise}
    \includegraphics[width=6cm]{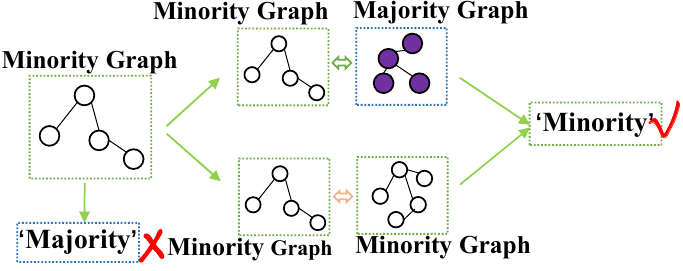}}
    \subfigure[]{\label{motivation-sub}
    \includegraphics[width=2.5cm]{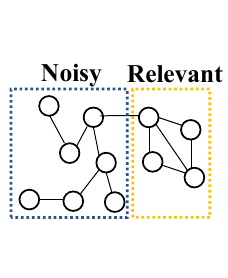}} 
    \caption{Motivation of pairwise- and subgraph-scale oversampling. (a) Pairwise-scale classification. A minority graph wrongly classified based on graph-scale information can be correctly classified based on graph interactions; (b) Subgraph-scale classification. In some graphs, only their subgraphs are relevant to the classification; the other parts are noisy. Oversampling the whole graphs may lead to the inclusion of more noise.}
    \label{motivation}
    \vspace{-0.5cm}
\end{figure}

In this paper, we propose a novel \underline{m}ulti-scale \underline{o}ver\underline{s}ampling GNN, namely MOSGNN, to learn various discriminative semantics within and between minority graphs for imbalanced graph classification. MOSGNN is specifically designed to leverage graph information at multiple scales, \ie, subgraph, graph, and pairwise-graph scales, to generate a large number of minority-graph-oriented samples of diversified semantic for mitigating the bias toward the majority graphs at different granularities in GNN-based models. 
To this end, in addition to the oversampling at the graph scale in general oversampling-based approaches, MOSGNN jointly optimizes two auxiliary objectives on pairwise graph relation prediction and multiple-instance-learning-based (MIL) subgraph classification.

The pairwise graph relation prediction is designed to extend the classification information of the minority graphs by exploiting the inter-graph interactions. Through pairing with other graphs, large-scale minority-oriented graph pairs can be generated easily, \ie, majority-minority and minority-minority pairs vs. majority-majority pairs, from which MOSGNN can learn discriminative pairwise patterns of the minority graphs against that of majority graph pairs. This enables MOSGNN to correctly classify minority graphs that are otherwise wrongly classified by the graph-scale oversampling module (see Fig.~\ref{motivation-pairwise}).
On the other hand, considering the specific structure characteristic of graphs and the fact that given a graph, only part of the graphs, \eg, a few subgraphs, are relevant to the graph classification~\cite{ma2022deep}
(Fig.~\ref{motivation-sub}), we design a subgraph-level oversampling to generate diverse subgraphs to represent graphs with different local semantics. Since one graph can have relevant and noisy subgraph patterns, we use MIL to allow MOSGNN to focus on only the relevant subgraphs of each graph. In doing so, MOSGNN learns expressive minority graph representations based on the rich semantics embedded within and between the graphs, enabling better classification performance.

In summary, we make three main contributions:
\begin{itemize}
    \item We introduce a novel deep multi-scale oversampling framework to address the imbalanced graph classification problem (Sec.~\ref{sec:framework}). It learns discriminative representations of the minority graphs with the support of oversampling the minority graphs that have largely augmented semantics at the subgraph, graph, and pairwise inter-graph levels. This is the first work that takes account of both within and between graph information to learn graph representations for imbalanced graph classification.
    \item The framework is instantiated into a novel model, called MOSGNN. The model is specifically designed to learn the minority graphs at multiple scales using graph convolutional networks jointly optimized with subgraph-level MIL, graph-level classification, and pairwise graph relation prediction (Sec. \ref{sec:model}). The two auxiliary tasks offer much stronger inductive knowledge than the current graph augmentation operation-based methods.
    \item Comprehensive experiments on 16 imbalanced graph datasets with various imbalanced ratios show that MOSGNN (i) significantly outperforms five competing methods (Sec.~\ref{subsec:sota})
    , and (ii) offers a generic framework, in which different advanced imbalanced learning loss functions
    and GNN backbones can be easily plugged in and obtain significantly improved classification performance (Secs.~\ref{subsec:loss} and \ref{subsec:other GNN}).
\end{itemize}

\section{Related Work}
\subsection{Imbalanced Learning}
Deep imbalanced learning is one of the most active research areas in recent years~\cite{zhang2021deep}. Class re-sampling, which seeks to balance the number of samples from different classes during training, is one of the most widely used types of method~\cite{he2009learning,chawla2002smote, kang2019decoupling,zang2021fasa}. 
Another popular approach is to re-balance classes through adjusting the loss values for the samples of different classes~\cite{lin2017focal,tan2020equalization,cao2019learning,cui2019class,ren2020balanced}. Some works also try to solve the problem by a post-hoc shifting of the model logits based on label frequencies~\cite{menon2020long,kini2021label,hong2021disentangling}. 
Instead of shift on the logits, \cite{du2021graph} puts forward class-imbalance learning based on the numerical label rather than the original logical label. Information augmentation addresses the imbalanced distribution through introducing additional data during training~\cite{zhang2017mixup,verma2019manifold,yun2019cutmix,chou2020remix}.
Unlike these methods, our work considers the unique properties of graphs and incorporates intra- and inter-graph information to learn multi-scale oversampling GNNs for imbalanced graph classification.


\subsection{Deep Imbalanced Graph Classification}
Most current deep methods for imbalanced classification in the graph domain focuses on \textit{node-level tasks}~\cite{liu2023survey,shi2020multi,wang2020network,zhao2021graphsmote,qu2021imgagn,park2021graphens,wang2021mixup,wang2021tackling}. 
Deep methods for graph-level imbalanced classification are very limited. There is one 
method dedicated to learn imbalanced graph structure information~\cite{liu2022size}, but it is size oriented. Graph-of-Graph Neural Networks (G$^2$GNN)~\cite{wang2022imbalanced} are designed and trained using augmented graphs to learn graph representations for imbalanced graph classification. However, the oversampling operation does not result in much extra semantic information for the minority graphs. Another shortcoming of G$^2$GNN is that the construction of GoG relies on kernel similarity among graphs, which is computationally costly for large-scale graph datasets.

\section{Framework}\label{sec:framework}
\subsection{Problem Statement}\label{subsec:problem}
This work tackles the problem of binary imbalanced graph classification. Specifically, given a set of $M$ graphs $\mathcal{G}=\{G_1, ..., G_M\}$ from a majority class and $N$ graphs $\mathcal{G^{\prime}}=\{G_{M+1}, ..., G_{M+N}\}$ from an under-represented class, where $N\ll M$, we aim at learning a mapping function $f\colon \mathcal{G}\cup\mathcal{G^{\prime}} \rightarrow\mathbb{R}$, parameterized by $\Theta$, such that $f(\hat{G}_i;\Theta)>f(\hat{G}_j;\Theta)$ if $\hat{G}_i\in\mathcal{G^{\prime}}$ and $\hat{G}_j\in\mathcal{G}$. 

\begin{figure}[h]
  \centering
  \includegraphics[width=9.2cm]{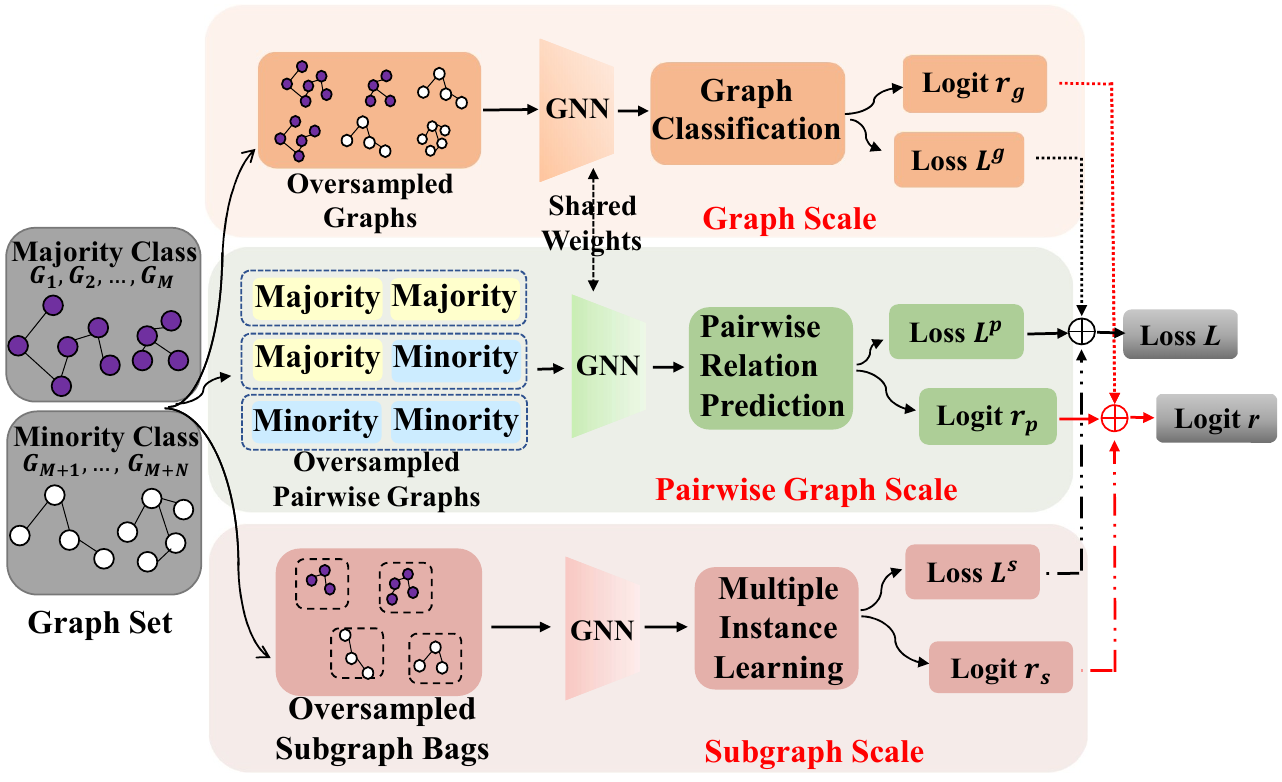}\hspace{-6pt}
  \caption{Overview of the proposed framework. It augments and trains a GNN model with oversampled graph data at the subgraph, graph, and pairwise inter-graph levels, to capture diversified intra- and inter-graph information for the classification of minority graphs. To achieve this goal, two auxiliary objectives, \ie, pairwise graph relation prediction and subgraph-based MIL, are combined with the standard graph classification objective to jointly optimize the GNN.}
  \label{fig:framework}
  \vspace{-0.2cm}
\end{figure}

\subsection{The Proposed Framework}\label{subsec:framework}
To solve the imbalanced graph classification problem, we propose a novel deep multi-scale oversampling framework that augments GNNs with oversampling at three different scales, \ie, subgraph, graph, and pairwise graph relation, to learn discriminative representations of the minority graphs. The key idea here is to learn graph representations from multiple-scale oversampling to capture diverse discriminative information of the minority class graphs. To this end, we develop a framework with three classification objectives from the perspectives of subgraph, graph and pairwise inter-graph scales, respectively. The objective of each classification branch helps achieve an oversampling of a particular scale. The final classification model is based on the combined results of three branches that are jointly optimized. Particularly, the overall objective of our framework can be generally formulated as:
\begin{equation}
    \mathcal{L}=L^g +\lambda L^p + \beta L^s,
\end{equation}
where $L^g$, $L^p$ and $L^s$ are the loss functions for enforcing oversampling from the graphs, pairwise graphs, and subgraphs respectively and $\lambda$ and $\beta$ are hyperparameters that balance the importance of the three scales.

The overall procedure of MOSGNN is illustrated in Figure~\ref{fig:framework}, which consists of the following three components:
\begin{itemize}
    \item \textit{Graph-level oversampling via standard graph classification}. In this component, a standard oversampling method is applied to $\mathcal{G^{\prime}}$. The oversampled minority graphs are then combined with $\mathcal{G}$ and used to train a GNN-based classifier with a loss function $L^g$, yielding a classification logit $\mathbf{r}_{g}$ for each graph. 
    \item \textit{Pairwise-graph oversampling via pairwise relation prediction}. This module aims to utilize the relation between every pair of graphs from $\mathcal{G}$ or $\mathcal{G^{\prime}}$ to gain discriminative inter-graph-level information for the minority class. The graph pairs can be majority-majority, majority-minority, and minority-minority graph combinations. The relation prediction is to discriminate the majority-majority pairs against the two types of minority-graph-oriented pairs via a loss function $L^p$. The oversampling is applied at the pairwise graph level such that we have a balanced set of majority-majority graph pairs and the majority-minority or minority-minority graph pairs. The relation prediction will output a prediction logit $\mathbf{r}_{p}$ for each graph pair.
    \item \textit{Subgraph-level oversampling via MIL}. This component is focused to apply the oversampling to the subgraph level. It assumes that, given a graph, only a few subgraphs within this graph are relevant to the graph classification. To accommodate this idea, we transform the graph classification into a weakly-supervised subgraph classification where a graph is represented by a bag of subgraphs and graph-level labels are given to perform the classification of the subgraphs. A loss function of top-$k$ multiple instance learning (MIL), $L^s$, is used to achieve this goal. Oversampling is applied to the minority class samples $\mathcal{G^{\prime}}$ before the generation of the subgraphs of each graph. Since $L^s$ is enforced on the subgraph samples, the oversampling is equivalent to a subgraph oversampling. This component will combine the prediction logits of subgraph bags to yield a graph-level prediction score $\mathbf{r}_s$ for each graph.
\end{itemize}

The three loss functions $L^g$, $L^p$, and $L^s$ are jointly minimized at the training stage. During inference, three prediction logits $\mathbf{r}_{g}$, $\mathbf{r}_{p}$ and $\mathbf{r}_s$ 
are lastly combined to define the prediction probability of a test graph $\hat{G}$ as
\begin{equation}
    \mathbf{r}(\hat{G}) =  \mathbf{r}_{g} +\lambda \mathbf{r}_{p} + \beta \mathbf{r}_s.
\end{equation}

\section{Model} \label{sec:model}
This section presents an instantiation of our framework, namely MOSGNN. Various GNNs and pooling operations 
can be used as the graph representation learning backbone. In our model, the commonly used GCN with the recently proposed MVPool~\cite{zhang2021hierarchical} and shared weight parameters is used by default in all three branches.

\subsection{Graph-level Oversampling}
A general graph-level oversampling is used to address the class imbalance in the learning of the graph representations via GNNs. 
Specifically, this branch feeds the balanced graph embeddings obtained from a GCN backbone into a multi-layer perception (MLP) classifier to calculate the probability of each graph belonging to each class as follows:
\begin{equation}\label{s^g_i}
	\mathbf{s}^{g}_i = MLP^g(\mathbf{h}_i; \theta_g),
\end{equation}
where $\mathbf{h}_i=GCN(G_i; \theta_{gcn})$ is the embedding of a graph $G_i$ from the majority graph set $\mathcal{G}$ or the oversampled minority graph set $\mathcal{G^{\prime}}$.
To optimize the GCN and MLP layers, a standard cross-entropy loss function is used:
\begin{equation}\label{eqn:graph}
	L^{g}=\sum_{i=1}^{2M} CE(\mathbf{s}^{g}_i, l_i),
\end{equation}
where $\{l_i\}^{2M}_{i=1}$ is the label set and $CE(\cdot)$ is the cross entropy loss function. 
This standard loss enforces the MOSGNN model to learn discriminative representations from the full graphs. The two auxiliary loss functions below complement this loss function to learn more discriminative graph representations from the inter-graph relation and subgraph perspectives.

\subsection{Pairwise-graph-level Oversampling}\label{subsec:pairwise}
The pairwise graph oversampling is devised to extend the classification information of the minority graphs by exploiting the pairwise inter-graph interactions. To this end, a self-supervised learning task is designed to reformulate the primary classification problem as a pairwise relation prediction task. In particular, we first pair the majority and minority graphs into three types of graph pairs, including majority-majority, majority-minority, and minority-minority pairs, creating a new pairwise graph set $\mathcal{P}=\{(\mathbf{h}_i,\mathbf{h}_j), y_{ij}|\mathbf{h}_i,\mathbf{h}_j\in\mathcal{H}\}$, where $\mathcal{H}$ is the graph embedding set produced by a GCN backbone, $(\mathbf{h}_i,\mathbf{h}_j)$ is a graph pair randomly sampled from $\mathcal{G}$ or oversampled $\mathcal{G^{\prime}}$, and $y_{ij}$ is a surrogate label of the pair $(\mathbf{h}_i,\mathbf{h}_j)$. Each pair $(\mathbf{h}_i,\mathbf{h}_j)$ has one of the three pairwise relations: $P_{maj,maj}$, $P_{maj,min}$ and $P_{min,min}$. To significantly extend the minority class data and balance the pairwise graph samples, the objective is further reformulated as a new binary classification task, aiming at discriminating majority graph pairs $P_{maj,maj}$ from minority-graph-related pairs $P_{maj,min}$/$P_{min,min}$ instead. In doing so, we leverage the interactions within the minority graphs and between the minority and majority graphs to generate a substantially large number of minority-graph-oriented pair samples. Learning to differentiate $P_{maj,maj}$ from $P_{maj,min}$/$P_{min,min}$ would then capture discriminative information of the minority graphs at the inter-graph level. More specifically, the pairwise graph learning is formally defined as:
\begin{equation}\label{eqn:graphpair}
	L^{p}=\sum\nolimits_{(i,j)\in \mathcal{P}}CE(\mathbf{s}^{p}_{ij}, y_{ij}),
\end{equation}
where $CE(\cdot)$ is the cross entropy loss function, $\mathbf{s}^{p}_{ij} = \mathcal{R}^p(\mathbf{h}_i, \mathbf{h}_j;\theta_p)$ and $\mathbf{h}_i=GCN(G_i; \theta_{gcn})$ (the same GCN is used to obtain $\mathbf{h}_j$), $\mathcal{R}^p(\cdot,\cdot;\theta)$ is the relation learning function, and $y_{ij}$ is the relation label of pair $(\mathbf{h}_i, \mathbf{h}_j)$. Since the intrinsic relationship between samples can be diverse across the datasets, it is difficult to explicitly define it. In MOSGNN, we employ a minority graph prediction network to learn it as:
\begin{equation}\label{eqn:graphpairMLP}
\mathbf{s}^{p}_{ij} = MLP^p(\mathbf{h}_i\odot\mathbf{h}_j;\theta_p),
\end{equation}
where 
$\odot$ is the concatenate operation and $y_{ij}=0$ if $(\mathbf{h}_i,\mathbf{h}_j)$ is a $P_{maj,maj}$ pair and $y_{ij}=1$ otherwise.

\subsection{Subgraph-level Oversampling}
Graph have rich local information, \eg, subgraph information can support the accurate classification of graphs, while using full graph may fail due to the presence of possible noisy nodes or subgraphs in the graphs. Considering this, we further devise a subgraph-level oversampling to improve the graph classification results. This equals to a weakly-supervised classification task where graph-level labels are given to perform the classification of subgraphs, with each graph represented by a bag of subgraphs. Thus, we employ a widely-used weakly-supervised learning technique MIL to implement this component. 

Specifically, we first randomly remove some nodes and edges to generate a bag of $q$ subgraphs $\{G^{n}_i\}_{n=1}^q$ for each graph $G_i$. A GCN with parameter $\theta_{sgcn}$ is then applied to each subgraph bag to gain their embeddings $\mathcal{B}_{G_i} = \{\mathbf{h}^n_i\}_{n=1}^q$, where $\mathbf{h}^n_i$ is the embedding of subgraph $G^{n}_i$. Considering the small number of minority data, we also pair the embedding bags, \ie, majority-majority, majority-minority and minority-minority subgraph pairs to produce more minority data. Similar to Sec. \ref{subsec:pairwise}, the majority-majority pair bags are regarded as one class and the other two types of pairs are treated as another class. That is, the subgraph bags are paired with each other, resulting in 
$\mathcal{B}=\{(\mathcal{B}_{G_i}, \mathcal{B}_{G_j}),y_{ij}^s| G_i, G_j\in\mathcal{G}\cup\mathcal{G^{\prime}}\}$, where $y_{ij}^s=0$ if $(\mathcal{B}_{G_i}, \mathcal{B}_{G_j})$ is a $P_{maj,maj}$ pair and $y_{ij}^s=1$ otherwise. The pairwised subgraph bags focuses on discriminative substructures while the graphs in pairwise level are centered on interaction between graphs. They can find similar patterns, but they also complement to each other. 

In implementing the MIL task, we employ the recently proposed feature magnitude learning-based top-$k$ MIL approach \cite{tian2021weakly}. Specifically, given a pair of subgraph bags $(\mathcal{B}_{G_i}, \mathcal{B}_{G_j})$, the top-$k$ MIL learning is implemented to map the pair to a prediction score by:
\begin{equation}\label{eqn:sub_s}
	 \mathbf{s}_{ij}^s = MIL((\mathcal{B}_{G_i},\mathcal{B}_{G_j}); \theta_s)=mean(\mathbf{S}_{ij}^{k}; \theta_s),
\end{equation}
where $\mathbf{S}_{ij}^k = topk(\{\mathbf{s}_i^n\}_{n=1}^q \cup \{\mathbf{s}_j^n\}_{n=1}^q$) are the top-$k$ prediction scores of the predictions from both subgraph bags $\mathcal{B}_{G_i}$ and $\mathcal{B}_{G_j}$, and $s_i^n= MLP^s(\mathbf{h}^n_i; \theta_s)$ is the prediction score of a single subgraph $\mathbf{h}^n_i\in \mathcal{B}_{G_i}$. Note that $MLP^s$ takes individual bags of subgraph embeddings, rather than the concatenated embeddings of subgraphs to avoid prohibitive computational cost, since there are a large number of possible subgraph pairs per two graphs. The pairwise subgraph interaction is captured instead at the output layer via the top-$k$ prediction across the two subgraph bags. 
Then, the binary cross-entropy loss is applied to optimize the MIL model: 
\begin{equation}
	L_{mil}= \sum\nolimits_{(i,j)\in \mathcal{B}} CE(\mathbf{s}_{ij}^s,y_{ij}^s).
\end{equation}

In \cite{tian2021weakly}, it is shown that enforcing a feature magnitude-based gap between the classes at the feature representation layers can further improve the classification performance. We accordingly employ this regularization here as:
\begin{equation}
	L_{reg} \!=\! \sum\nolimits_{(i,j)\in \mathcal{B}} \mathrm{max}\!\left(0,\! m\!-\!g_{k}(\mathcal{B}_{G_i})+g_k(\mathcal{B}_{G_j})\right), 
\end{equation}
where $G_i$ is a minority class graph, $G_j$ is a majority class graph, $m$ is a pre-defined margin, and $g_{k}(\cdot)$ denotes the mean of the top-$k$ $L_2$-norm values across the subgraph embeddings in each bag. Thus, the overall loss function of the subgraph-scale branch is as:
\begin{equation}\label{eqn:subgraph}
	L^{s}=L_{mil} +\eta L_{reg},
\end{equation}
where $\eta$ is a parameter that balances the two loss terms.

\subsection{Training and Inference}
At the training stage, we first use oversampling and GCNs (with MVPool-based graph pooling) to generate class-balanced batches of graph/subgraph embeddings in each branch, 
and then jointly optimize the following overall loss:
\begin{equation}\label{eqn:mos}
	\mathcal{L}=\underbrace{L^g}_{graphs} +\underbrace{\lambda L^p}_{pairwise\;graphs} + \underbrace{\beta L^s}_{subgraphs},
\end{equation}
where $L^g$, $L^p$, and $L^s$ are respectively specified as \eqref{eqn:graph}, \eqref{eqn:graphpair}, and \eqref{eqn:subgraph}, and $\lambda$ and $\beta$ are hyperparameters that balance the influence of the three components on the overall classification. It yields a MOSGNN model with optimal parameters $\Theta^*=\{\theta_{gcn}^*, \theta_{sgcn}^*, \theta_g^*, \theta_p^*, \theta_s^*\}$.

\begin{table*}[htbp]
\caption{F1 score (mean$\pm$std) of MOSGNN and five SOTA competing methods.
$\#pos$ and $\#total$ denote the number of graph samples in the minority class and in the full dataset, respectively. Ratio denotes the ratio of the majority class size to the minority class. The best performance per dataset is boldfaced.}
\centering
\label{f1}
\setlength{\tabcolsep}{0.6mm}
\scalebox{0.8}{
\begin{tabular}{lrrr|cccccc}
\hline\hline
{\textbf{Dataset}} & {\textbf{$\#pos$}} &{\textbf{$\#total$}}&{\textbf{Ratio}} &  
\textbf{SMOTE} & \textbf{Mixup} & \textbf{Oversampling} & \textbf{FocalLoss} & \textbf{LALoss} & \textbf{MOSGNN} \\
\hline
NCI1 & 1793 & 37349 & $19.8\colon1$ & $0.2531\pm0.0299$ & $0.3837\pm0.0376$ & $0.4470\pm0.0369$ & $0.4250\pm0.0330$ & $0.3712\pm0.0028$ & $\mathbf{0.4772\pm0.0230}$\\
NCI33 & 1467 & 37022 & $24.2\colon1$ & $0.2083\pm0.0135$ & $0.3362\pm0.0233$ & $0.4394\pm0.0197$ & $0.4153\pm0.0155$ & $0.3179\pm0.0398$ & $\mathbf{0.4684\pm0.0108}$ \\
NCI41 & 1350 & 25336 & $17.8\colon1$  & $0.2516\pm0.0212$ & $0.3816\pm0.0395$ & $0.4315\pm0.0214$ & $0.4154\pm0.0097$ & $0.3016\pm0.0049$ & $\mathbf{0.4823\pm0.0151}$\\
NCI47 & 1735 & 37298 & $20.5\colon1$ & $0.2628\pm0.0024$ & $0.3162\pm0.0317$ & $0.4276\pm0.0089$ & $0.3829\pm0.0217$ & $0.3240\pm0.0030$ & $\mathbf{0.4477\pm0.0188}$\\
NCI81 & 2081 & 37549 & $17.0\colon1$ & $0.2738\pm0.0294$ & $0.3585\pm0.0331$ & $0.4370\pm0.0187$ & $0.4282\pm0.0256$ & $0.3799\pm0.0094$ & $\mathbf{0.4814\pm0.0132}$\\
NCI83 & 1959 & 25550 & $12.0\colon1$ & $0.2875\pm0.0086$ & $0.3403\pm0.0346$ & $0.4124\pm0.0216$ & $0.4281\pm0.0022$ & $0.3543\pm0.0149$ & $\mathbf{0.4399\pm0.0135}$\\
NCI109 & 1773 & 37518 & $20.2\colon1$ & $0.2643\pm0.0229$ & $0.3472\pm0.0359$ & $0.4235\pm0.0064$ & $0.3892\pm0.0111$ & $0.3388\pm0.0087$ & $\mathbf{0.4661\pm0.0200}$\\
NCI123 & 2715 & 36903 & $12.6\colon1$ & $0.2696\pm0.0246$ & $0.2909\pm0.0819$ & $0.3890\pm0.0169$ & $0.4004\pm0.0107$ & $0.3516\pm0.0087$ & $\mathbf{0.4263\pm0.0128}$\\
NCI145 & 1641 & 37043 & $21.6\colon1$ & $0.2721\pm0.0217$ & $0.3195\pm0.0370$ & $0.4512\pm0.0009$ & $0.4086\pm0.0279$ & $0.3346\pm0.0148$ & $\mathbf{0.4684\pm0.0180}$\\
BZR & 86 &405  & $3.7\colon1$ & $0.5203\pm0.0617$ & $0.4854\pm0.0328$ & $0.5416\pm0.0299$ & $0.4591\pm0.0519$ & $0.5024\pm0.0321$ & $\mathbf{0.5890\pm0.0249}$\\
COX2 & 102 & 467 & $3.6\colon1$ & $0.3276\pm0.0475$ & $0.2825\pm0.2007$ & $0.4380\pm0.0571$ & $0.3426\pm0.2445$ & $0.0889\pm0.1257$ & $\mathbf{0.4938\pm0.0545}$\\
P388 & 2298 & 41472 & $17.1\colon1$ & $0.3226\pm0.0224$ & $0.5223\pm0.0036$ & $0.5471\pm0.0134$ & $0.5319\pm0.0164$ & $0.4646\pm0.0094$ & $\mathbf{0.5606\pm0.0134}$\\
Aromatase & 360 & 7226 &  $19.1\colon1$ &$0.0444\pm0.0629$ & $0.0889\pm0.0685$ & $0.2222\pm0.0856$ & $0.1005\pm0.0741$ & $0.1721\pm0.0582$ & $\mathbf{0.2820\pm0.0910}$\\
ATAD5 & 338 & 9091 & $25.9\colon1$ & $0.0614\pm0.0482$ & $0.0000\pm0.0000$ & $0.1197\pm0.0579$ & $0.1393\pm0.0550$ & $0.1455\pm0.0157$ & $\mathbf{0.2231\pm0.0271}$\\
ER & 937 & 	7697 & $7.2\colon1$ &$0.0490\pm0.0416$ & $0.1237\pm0.0494$ & $0.1948\pm0.0277$ & $0.1373\pm0.0480$ & $0.0875\pm0.0178$ & $\mathbf{0.2014\pm0.0226}$\\
p53 & 537 & 8634 & $15.1\colon1$ &$0.2296\pm0.0469$ & $0.1618\pm0.1145$ & $0.2109\pm0.0264$ & $0.2550\pm0.0290$ & $0.2202\pm0.0236$ & $\mathbf{0.2580\pm0.0495}$\\
\hline
\multicolumn{4}{r|}{\textbf{Average Rank}} &  5.4 &  4.8& 2.4&  3.1 & 4.3   & 1 \\
\multicolumn{4}{r|}{\textbf{p-value}} &  0.0004 &  0.0004& 0.0004&  0.0004 & 0.0004   & - \\
\hline\hline
\end{tabular}
}
\vspace{-0.2cm}
\end{table*}

During inference, given a test graph sample $G$, its graph embedding should be paired with other graph embeddings to extract relationships between them. We evaluated the pair of the test graph with itself and graphs randomly sampled from the training data and obtained similar results. Therefore, we choose to easily pair test graphs with themselves to obtain the prediction of the pairwise scale. Finally, the classification probability predicted by MOSGNN for $G$ is defined as:
\begin{equation}
		\mathbf{r}\!(\!G\!)\!\!= \!\!M\!L\!P^g\!(\mathbf{h};\!\theta_g^*)\!+\!\lambda \!M\!L\!P^p\!(\!\mathbf{h}\odot\mathbf{h};\!\theta_p^*\!)\! + \!\beta \!M\!I\!L\!(\!\mathcal{B}_{G},\!\theta_s^*\!),
\end{equation}
where $\mathbf{h}\!=\!GCN(G,\theta_{gcn}^*)$. Class with maximum probability in $\mathbf{r}(G)$ is the class of $G$ predicted by MOSGNN.

\subsection{Enabling Other Imbalanced Learning Loss Functions}
There have been many imbalanced learning loss functions adaptively assigning larger weights to the minority samples during training.
We found empirically that the performance of these advanced loss functions can be further improved under our MOSGNN framework by simply plugging these losses to replace the cross entropy loss in MOSGNN (see Sec.~\ref{subsec:loss}).

\subsection{Complexity Analysis}
Next, we analyze the time complexity of our proposed MOSGNN. The graph-scale oversampling is the vanilla oversampling-based GNN whose computational requirements come from the oversampling operation, GNN and the MLP classifier. Our oversampling operation is simply duplicating data and its complexity is $O(M)$. Since the GNN encoder can be various and we use $O_{GNN}$ to stand for its complexity. The MLP classifier whose layer number and dimension are constant requires $O(M)$ time. The extra computational requirements are due to the pairwise-graph-scale and subgraph-scale modules. As for the pairwise graph scale, the oversampling and GNN operations can use the results from the graph scale. Both the graph pairing and relation prediction steps require $O(M)$. Therefore, the pairwise graph scale module needs $O(M)$ additional time. As in \cite{wang2022imbalanced}, the complexity of subgraph augmentation is linearly proportional to the size of graph and imposes no additional time compared with GNN encoder. The GNN operation needs extra $O_{GNN}$ time, while the MIL module takes an extra $O(M)$ time for score calculation and $O(kM\log q)\approx O(M)$ for the selection of top-$k$ scores in each bag. Therefore, the total time complexity of our MOSGNN is $O(M)+O_{GNN}$. For example, with GCN as GNN backbone, the time complexity will be $O(M(N_e+1))$, where $N_e$ denotes the maximal number of edges in graphs of $\mathcal{G}$ and $\mathcal{G}^{'}$.

\section{Experiments and Results}
\subsection{Datasets}
16 publicly available datasets with different imbalanced ratios are used to evaluate the effectiveness of our MOSGNN, including 9 NCI chemical compound graph datasets for anticancer activity prediction from GRAND-Lab\footnote{https://github.com/GRAND-Lab/graph\_datasets} and 7 datasets from the TUDataset graph classification benchmark \cite{Morris+2020}. 

\subsection{Competing Methods and Evaluation Metric}
There are three exisiting GNN-based imbalanced graph classification methods, \ie, mixup~\cite{wang2021mixup}, G$^2$GNN~\cite{wang2022imbalanced} and SOLT-GNN\cite{liu2022size}. The construction of GoG in G$^2$GNN relies on GraKeL~\cite{JMLR:v21:18-370}, which supports only small datasets. Since most of our datasets have a large scale (NCI and Tox datasets have $30$K and $8$K samples) and they are not supported by GraKeL, we cannot apply G$^2$GNN to these datasets. Besides, the imbalance issue in SOLT-GNN is on the graph size, rather than the class imbalance, which is different from our setting. Therefore, our MOSGNN is compared to five state-of-the-art (SOTA) competing methods from four different directions: i) \textbf{Re-sampling:} Oversampling \cite{he2009learning} and SMOTE ~\cite{chawla2002smote}; ii) \textbf{Data Augmentation:} Mixup ~\cite{wang2021mixup}; iii) \textbf{Re-weighting:} FocalLoss~\cite{lin2017focal}; and iv) \textbf{Logit Adjustment:} LALoss ~\cite{menon2020long}.



Since the performance of accurately classifying minority samples is typically the key focus, we employ F1 score on the minority class as the primary evaluation metric. Higher F1 score indicates better classification performance. 
We report the mean results and standard deviation based on 3-fold cross-validation for all datasets. 


\subsection{Implementation Details}\label{subsec: setting}
The following parameters are set by default for MOSGNN and its competing methods on all 16 datasets: the batch size is set to 256, the number of GCN layers combined with MVPool is 3, the dimension of hidden layer in the MVPool-based graph pooling is 128, the pooling ratio of MVPool is 0.8, and the number of epochs is 200. The learning rate is $10^{-3}$ for all method except LALoss, which use a linear learning rate warming up in the first five epochs to reach the base learning rate $0.1$ as recommended in \cite{menon2020long}. All methods are trained with Adam except LALoss, which is trained with the SGD optimizer following \cite{menon2020long}. All methods select the best decision threshold with the classification probability varying from $0.3$ to $0.9$ using a step size of $0.1$. The Mixup hyperparameter used to combine two graphs into one is selected from $\{0.05,0.1,0.15,0.2,0.25,0.3\}$. The parameters $\lambda$ and $\beta$ in MOSGNN are selected from $\{1.0, 0.5, 0.25, 0.0\}$. The parameter $q$ in the subgraph branch of MOSGNN is set to $10$, while the other three parameters $m$, $k$ and $\eta$ are set to $100$, $3$ and $0.0001$ respectively, as suggested in \cite{tian2021weakly}. Other parameters in the competitors are set as suggested in their original works.

\begin{table}[htbp]
\vspace{-0.2cm}
\centering
\caption{F1 score (mean) using MOSGNN-enabled FocalLoss and LALoss, with the original FocalLoss and LALoss as baselines. `Diff.' denotes the F1 score improvement ($\textcolor{magenta}\uparrow$) or decrease ($\textcolor{blue}\downarrow$) of `MOS-X' compared to the original `X' loss.}
\label{difloss-f1}
\setlength{\tabcolsep}{0.2mm}
\scalebox{0.78}{
\begin{tabular}{l|ccr||ccr}
\hline\hline
{\textbf{Dataset}} &  \textbf{Focal} &  \textbf{MOS-Focal} & \textbf{Diff.} &  \textbf{LA} &  \textbf{MOS-LA} & \textbf{Diff.}\\
\hline
NCI1 & $0.4250$ & $0.4971$ & $0.0721$ $\textcolor{magenta}\uparrow$ & $0.3712$ & $0.4179$ & $0.0467$ $\textcolor{magenta}\uparrow$\\
NCI33 & $0.4153$ & $0.4558$ & $0.0405$ $\textcolor{magenta}\uparrow$ & $0.3179$ & $0.3962$ & $0.0783$ $\textcolor{magenta}\uparrow$ \\
NCI41 & $0.4154$ & $0.4503$ & $0.0349$ $\textcolor{magenta}\uparrow$ & $0.3016$ & $0.4211$ & $0.1195$ $\textcolor{magenta}\uparrow$\\
NCI47 & $0.3829$ & $0.4342$ & $0.0513$ $\textcolor{magenta}\uparrow$ & $0.3240$ & $0.3778$ & $0.0538$ $\textcolor{magenta}\uparrow$\\
NCI81 & $0.4282$ & $0.4859$ & $0.0577$ $\textcolor{magenta}\uparrow$ & $0.3799$ & $0.4193$ & $0.0394$ $\textcolor{magenta}\uparrow$\\
NCI83 & $0.4281$ & $0.4453$ & $0.0172$ $\textcolor{magenta}\uparrow$ & $0.3543$ & $0.4263$ & $0.0720$ $\textcolor{magenta}\uparrow$\\
NCI109 & $0.3892$ & $0.4788$ & $0.0896$ $\textcolor{magenta}\uparrow$ & $0.3388$ & $0.4302$ & $0.0914$ $\textcolor{magenta}\uparrow$\\
NCI123 & $0.4004$ & $0.4267$ & $0.0263$ $\textcolor{magenta}\uparrow$ & $0.3516$ & $0.3888$ & $0.0372$ $\textcolor{magenta}\uparrow$\\
NCI145 & $0.4086$ & $0.4604$ & $0.0518$ $\textcolor{magenta}\uparrow$ & $0.3346$ & $0.4105$ & $0.0759$ $\textcolor{magenta}\uparrow$\\
BZR & $0.4591$ & $0.5296$ & $0.0705$ $\textcolor{magenta}\uparrow$ & $0.5024$ & $0.5688$ & $0.0664$ $\textcolor{magenta}\uparrow$\\
COX2 & $0.3426$ & $0.4725$ & $0.1299$ $\textcolor{magenta}\uparrow$ & $0.0889$ & $0.4177$ & $0.3288$ $\textcolor{magenta}\uparrow$\\
P388 & $0.5319$ & $0.5628$ & $0.0309$ $\textcolor{magenta}\uparrow$ & $0.4646$ & $0.5379$ & $0.0733$ $\textcolor{magenta}\uparrow$\\
Aromatase & $0.1005$ & $0.1554$ & $0.0549$ $\textcolor{magenta}\uparrow$ & $0.1721$ & $0.2230$ & $0.0509$ $\textcolor{magenta}\uparrow$\\
ATAD5 & $0.1393$ & $0.1365$ & $-0.0028$ $\textcolor{blue}\downarrow$ & $0.1455$ & $0.1739$ & $0.0284$ $\textcolor{magenta}\uparrow$\\
ER & $0.1373$ & $0.2610$ & $0.1237$ $\textcolor{magenta}\uparrow$ & $0.0875$ & $0.1848$ & $0.0973$ $\textcolor{magenta}\uparrow$\\
p53 & $0.2550$ & $0.1536$ & $-0.1014$ $\textcolor{blue}\downarrow$ & $0.2202$ & $0.1736$ & $-0.0466$ $\textcolor{blue}\downarrow$\\
\hline
\textbf{p-value}  & 0.0061& - & &   0.0009&- &   \\
\hline\hline
\end{tabular}
}
\vspace{-0.3cm}
\end{table}

\subsection{Comparison to SOTA Models}\label{subsec:sota}
The results of comparing MOSGNN to five competing methods are reported in Table ~\ref{f1}. Our MOSGNN performs best on all datasets. The improvement of MOSGNN compared with the best competing method per dataset is large on most datasets, \eg, NCI33 ($2.9\%$), NCI109 ($4.3\%$), NCI81 ($4.4\%$), BZR ($4.7\%$), NCI41 ($5.1\%$), COX2 ($5.5\%$), Aromatase ($6\%$) and ATAD5 ($7.8\%$). Oversampling, FocalLoss and LALoss are the three most effective competing methods, but they cannot well leverage the intra- and inter-graph structure information, learning less discriminative representations.
MOSGNN achieves this by the multi-scale oversampling and largely outperforms them. The p-value results indicate that the improvement of MOSGNN over all the competitors is significant at the $99\%$ confidence level, demonstrating the superiority of MOSGNN on graph datasets with diverse imbalanced ratios. 


\subsection{Enabling Different Loss Functions}\label{subsec:loss}
This section evaluates the applicability of our model to enable other imbalanced learning loss functions. In detail, we replace the cross-entropy loss in MOSGNN with FocalLoss and LALoss, named as MOS-Focal and MOS-LA respectively. 
The results are reported in Table~\ref{difloss-f1}, showing that MOS-Focal and MOS-LA substantially outperform FocalLoss and LALoss respectively on all datasets except ATAD5 and p53 for FocalLoss and p53 for LALoss. The average improvement is $4.6\%$ and $7.6\%$ and the maximal improvement can be up to about $13\%$ and $33\%$ w.r.t. FocalLoss and LALoss respectively. The paired signed-rank test indicates the improvement across the 16 datasets is significant at 99\% confidence level. The performance drop of MOS-Focal w.r.t. FocalLoss on Aromatase are very marginal, having only a difference of $0.28\%$. The decreased performance on p53 is relatively large, since the subgraph-scale oversampling in MOSGNN works adversely on p53 (as shown in Table \ref{ablation-f1}) and this effect is further enlarged likely due to the sample weighting schemes in two losses.

\subsection{Enabling Different GNN Backbones}\label{subsec:other GNN}
This section measures the applicability of MOSGNN framework to other GNN backbones. We replace the GCN in our model with GIN~\cite{xu2018powerful} and GAT~\cite{velivckovic2017graph} respectively and use sum/mean pooling as the readout module. 
Since the Oversampling perform best in the competitors, we only compare Oversampling with our method in this experiment. The results are shown in Table~\ref{otherGNN}. It is obvious that the usage of GIN brings improvement on most datasets for both MOSGNN and Oversampling and our MOSGNN has better results on all datasets.
The enhancement brought by GAT is limited, which might be because of the usage of MVPool with GCN in original MOSGNN. However, our MOSGNN still has better performance than Oversampling on all datasets.

\begin{table}[htbp]
\vspace{-0.2cm}
\caption{F1 score results (mean$\pm$std) of using GIN/GAT as backbones. The best performance is boldfaced. `Baseline' denotes `Oversampling'.}
\label{otherGNN}
\centering
\setlength{\tabcolsep}{0.3mm}
\scalebox{0.78}{
\begin{tabular}{l|cc|cc}
\hline\hline
\multirow{2}{*}{\textbf{Dataset}} & \multicolumn{2}{c|}{\textbf{GIN}} & \multicolumn{2}{c}{\textbf{GAT}}\\
\cline{2-5}
&\multicolumn{1}{c}{\textbf{Baseline}} & \multicolumn{1}{c|}{\textbf{MOSGNN}}&\multicolumn{1}{c}{\textbf{Baseline}} & \multicolumn{1}{c}{\textbf{MOSGNN}}\\
\hline
NCI1&$0.541\pm0.006$&$\mathbf{0.542\pm0.033}$&$0.440\pm0.035$&$\mathbf{0.472\pm0.016}$\\
NCI33&$0.489\pm0.008$&$\mathbf{0.534\pm0.013}$&$0.417\pm0.022$&$\mathbf{0.456\pm0.022}$\\
NCI41&$0.509\pm0.007$&$\mathbf{0.537\pm0.018}$&$0.433\pm0.0075$&$\mathbf{0.458\pm0.004}$\\
NCI47&$0.498\pm0.026$&$\mathbf{0.512\pm0.004}$&$0.399\pm0.015$&$\mathbf{0.453\pm0.016}$\\
NCI81&$0.534\pm0.011$&$\mathbf{0.540\pm0.017}$&$0.459\pm0.019$&$\mathbf{0.474\pm0.021}$\\
NCI83&$0.489\pm0.015$&$\mathbf{0.504\pm0.007}$&$0.436\pm0.018$&$\mathbf{0.457\pm0.017}$\\
NCI109&$0.490\pm0.023$&$\mathbf{0.527\pm0.009}$&$0.438\pm0.023$&$\mathbf{0.442\pm0.015}$\\   
NCI123&$0.459\pm0.025$&$\mathbf{0.498\pm0.016}$&$0.417\pm0.015$&$\mathbf{0.431\pm0.016}$\\
NCI145&$0.536\pm0.009$&$\mathbf{0.545\pm0.010}$&$0.438\pm0.012$&$\mathbf{0.467\pm0.012}$\\
BZR&$0.503\pm0.044$&$\mathbf{0.586\pm0.017}$&$0.486\pm0.078$&$\mathbf{0.557\pm0.063}$\\
COX2&$0.443\pm0.055$&$\mathbf{0.502\pm0.040}$&$0.530\pm0.046$&$\mathbf{0.539\pm0.074}$\\ 
P388&$0.614\pm0.014$&$\mathbf{0.622\pm0.007}$&$0.547\pm0.013$&$\mathbf{0.566\pm0.008}$\\
Aromatase&$0.238\pm0.059$&$\mathbf{0.285\pm0.009}$&$0.236\pm0.027$&$\mathbf{0.307\pm0.030}$\\
ATAD5&$0.241\pm0.074$&$\mathbf{0.281\pm0.088}$&$0.187\pm0.093$&$\mathbf{0.231\pm0.047}$\\
ER&$0.165\pm0.026$&$\mathbf{0.207\pm0.037}$&$0.156\pm0.040$&$\mathbf{0.224\pm0.042}$\\
p53&$0.220\pm0.005$&$\mathbf{0.225\pm0.027}$&$0.205\pm0.077$&$\mathbf{0.229\pm0.030}$\\
\hline
\textbf{p-value} & 0.0004 & - & 0.0004 & - \\
\hline\hline
\end{tabular}
}
\vspace{-0.2cm}
\end{table}

\begin{figure}[htbp]
\centering
\includegraphics[width=8.5cm]{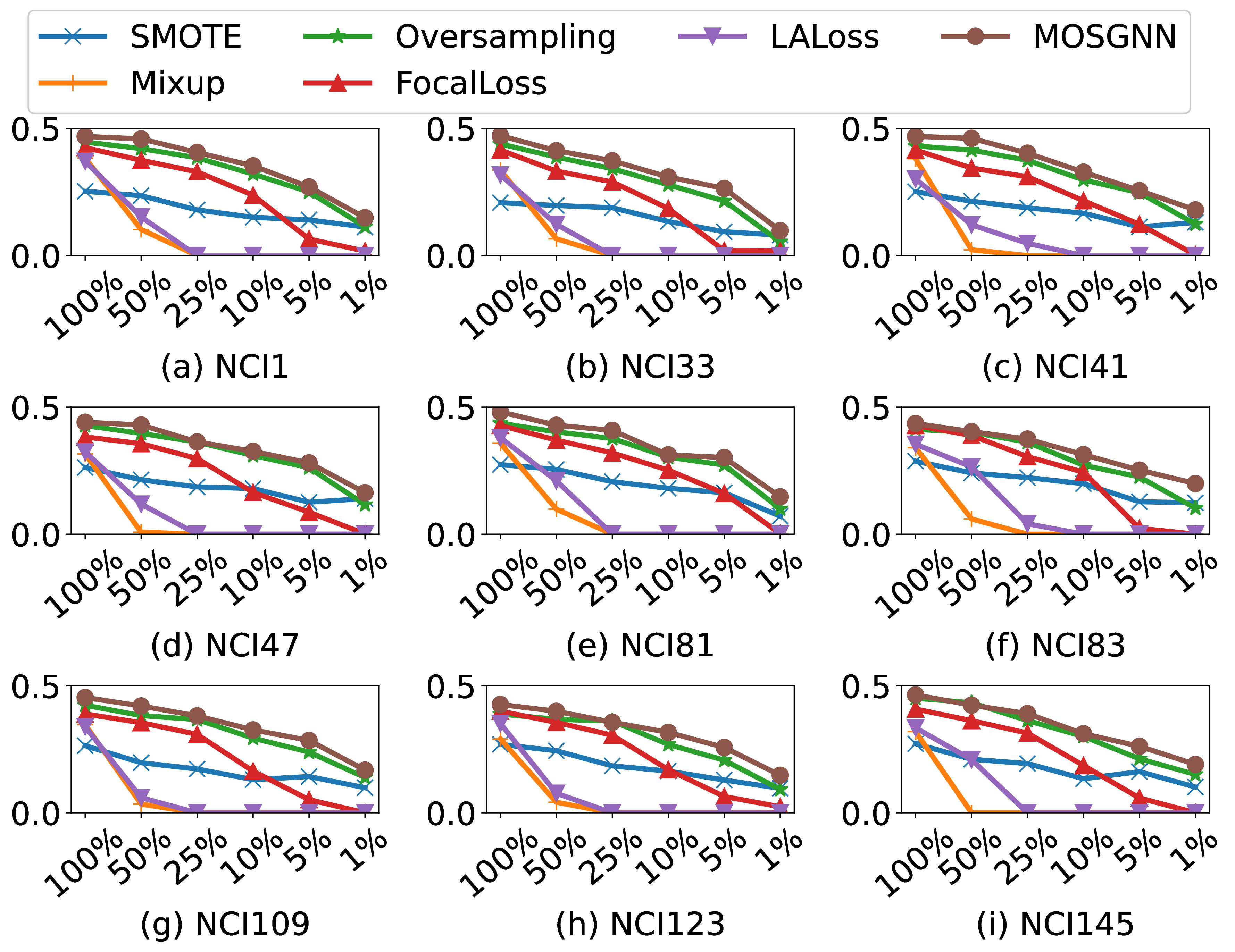}
\vspace{-0.2cm}
\caption{F1 score (y-axis) on nine NCI datasets with decreasing training data. }
\label{dataefficiency}
\end{figure}
\vspace{-0.2cm}

\subsection{Sample Efficiency}\label{subsec:data efficiency}
As the minority samples are typically difficult or costly to obtain, this section examines the performance of our model w.r.t. decreasing sample size of the minority class in training data, \ie, sample efficiency. 
We perform the experiment using $1\%$, $5\%$, $10\%$, $25\%$, $50\%$ and $100\%$ of training minority samples, respectively. 
The experiment is focused on the nine NCI datasets, with $\lambda=1$ and $\beta=1$ by default in MOSGNN; the competing methods use the hyperparameter settings that are optimal on the original training data per dataset. The F1 score results are displayed in Figure~\ref{dataefficiency}. MOSGNN obtains consistently higher F1 scores than the five competing methods in almost all cases. This superiority is particularly clear when only $5\%$ and $1\%$ minority samples are used. 
This might be because that MOSGNN is able to utilize the fine-grained subgraph information and the majority graph samples via MIL and pairwise relation prediction respectively, which helps alleviate the impacts of the decreasing minority data to some extent and maintain the consistent improvement.

\begin{table*}[htbp]
\caption{F1 score (mean$\pm$std) of MOSGNN and its ablated variants. The best performance per dataset is boldfaced.}
\centering
\label{ablation-f1}
\setlength{\tabcolsep}{0.6mm}
\scalebox{0.78}{
\begin{tabular}{l|ccccccc}
\hline\hline
{\textbf{Dataset}} &  
\textbf{$L^g$} & \textbf{$L^p$} & \textbf{$L^s$} & \textbf{$L^g$ \& $L^p$} & \textbf{$L^g$ \& $L^s$} & \textbf{$L^p$ \& $L^s$} & MOSGNN \\
\hline
NCI1 & $0.4470\pm0.0369$ & $0.4547\pm0.0096$ & $0.2147\pm0.1524$ & $0.4381\pm0.0277$ & $0.4572\pm0.0322$& $0.4684\pm0.0409$ & $\mathbf{0.4772\pm0.0230}$\\
NCI33 & $0.4394\pm0.0197$ & $0.4435\pm0.0233$ & $0.2090\pm0.1479$ & $0.4292\pm0.0057$ & $0.4410\pm0.0075$ &	$0.4384\pm0.0318$ & $\mathbf{0.4684\pm0.0108}$\\
NCI41 & $0.4315\pm0.0214$ & $0.4374\pm0.0381$ & $0.2831\pm0.0194$ & $0.4515\pm0.0164$ & $0.4379\pm0.0056$&$0.4608\pm 0.0268$ & $\mathbf{0.4823\pm0.0151}$\\
NCI47 & $0.4276\pm0.0089$ & $0.4022\pm0.0357$ & $0.1959\pm0.1386$ & $0.4257\pm0.0163$ & $0.3866\pm0.0433$&	$0.4334\pm0.0246$ & $\mathbf{0.4477\pm0.0188}$\\
NCI81 & $0.4370\pm0.0187$ & $0.4496\pm0.0106$ & $0.2062\pm0.1462$ & $0.4558\pm0.0168$ & $0.4397\pm0.0196$&	$0.4392\pm 0.0103$ & $\mathbf{0.4814\pm0.0132}$\\
NCI83 & $0.4124\pm0.0216$ & $0.4210\pm0.0287$ & $0.1834\pm0.1339$ & $\mathbf{0.4399\pm0.0135}$ & $0.4199\pm0.0111$&$0.4140\pm0.0139$ & $\mathbf{0.4399\pm0.0135}$\\
NCI109 & $0.4235\pm0.0064$ & $0.4261\pm0.0077$ & $0.2960\pm0.0451$ & $0.4579\pm0.0137$ & $0.4400\pm0.0081$&	$0.4445\pm0.0196$ & $\mathbf{0.4661\pm0.0200}$\\
NCI123 & $0.3890\pm0.0169$ & $0.4094\pm0.0079$ & $0.2955\pm0.0106$ & $0.4247\pm0.0152$ & $0.3909\pm0.0137$&	$0.4030\pm 0.0044$ & $\mathbf{0.4263\pm0.0128}$\\
NCI145 & $0.4512\pm0.0009$ & $0.4505\pm0.0150$ & $0.2009\pm0.1421$ & $0.4653\pm0.0114$ & $0.4470\pm0.0074$&	$0.4484\pm0.0126$ & $\mathbf{0.4684\pm0.0180}$\\
BZR & $0.5416\pm0.0299$ & $0.5759\pm0.0086$ & $0.3529\pm0.0476$ & $\mathbf{0.5890\pm0.0249}$ & $0.5150\pm 0.0250$&	$0.5826\pm 0.0448$ & $\mathbf{0.5890\pm0.0249}$\\
COX2 & $0.4380\pm0.0571$ & $0.4517\pm0.0579$ & $0.3533\pm0.0859$ & $0.4680\pm0.0767$ & $0.4046\pm0.0688$&$0.4555\pm0.0471$
 & $\mathbf{0.4938\pm0.0545}$\\
P388 & $0.5471\pm0.0134$ & $0.5512\pm0.0123$ & $0.2753\pm0.1947$ & $\mathbf{0.5606\pm0.0134}$ &  $0.5458\pm 0.0170$ & $0.5297\pm 0.0173$& $\mathbf{0.5606\pm0.0134}$\\
Aromatase & $0.2222\pm0.0856$ & $0.2594\pm0.0822$ & $0.0837\pm0.0595$ & $0.2428\pm0.0790$ &  $0.1248\pm0.0330$& $0.2061\pm0.0254$& $\mathbf{0.2820\pm0.0910}$\\
ATAD5 & $0.1197\pm0.0579$ & $0.2000\pm0.0636$ & $0.1707\pm0.0588$ & $\mathbf{0.2231\pm0.0271}$ & $0.1855\pm 0.0648$&	$0.1823\pm0.0908$& $\mathbf{0.2231\pm0.0271}$\\
ER & $0.1948\pm0.0277$ & $0.1518\pm0.0503$ & $0.0951\pm0.0380$ & $\mathbf{0.2014\pm0.0226}$ & $0.1734\pm0.0219 $ &  $0.1907\pm0.0525$& $\mathbf{0.2014\pm0.0226}$\\
p53 & $0.2109\pm0.0264$ & $\mathbf{0.2626\pm0.0277}$ & $0.1368\pm0.0282$ & $0.2580\pm0.0495$ & $ 0.2550\pm0.0260 $ & $ 0.2414\pm0.0262 $ & $0.2580\pm0.0495$\\
\hline
\textbf{Average Rank} &  4.9 & 3.6 & 6.9 & 2.4 & 4.7 & 4.0 & 1.1 \\
\textbf{p-value} &  0.0002&	0.0004&	0.0002&	0.0156&	0.0002&	0.0002
& - \\
\hline\hline
\end{tabular}
}
\vspace{-0.2cm}
\end{table*}

\subsection{Ablation Study}
This section examines the importance of three oversampling modules in our model. We first evaluate three oversampling branches separately ($L^g$, $L^p$, or $L^s$), and then incrementally add pairwise-graph-scale and subgraph-scale branches until we obtain the full model MOSGNN. The results are reported in Table~\ref{ablation-f1}. It is shown that using graph-scale ($L^g$) or pairwise-graph-scale oversampling ($L^p$) individually can achieve good performance.
Jointly optimizing $L^g$ and $L^p$ further improves the individual performance, indicating the complementary information gained from two types of oversampling. Using $L^s$ only cannot classify samples successfully, since the subgraph-based classification discards many nodes per graph, but it can still achieve fairly good F1 scores on some datasets, indicating that important information in some datasets are actually embedded in subgraphs.
Therefore, combining $L^s$ with $L^g$ or $L^p$ can obtain better performance than using single scale on those datasets. The worse performance obtained by combinations of two scales in $L^g$, $L^p$ and $L^s$ might be because that pairs of the three scales do not capture the full graph semantic well and part of their results are not consistent. This problem can be alleviated when three scales are used together, which is exactly what we have in our MOSGNN. Overall, the three oversampling done by $L^g$, $L^p$, and $L^s$ helps significantly enhance the representations of minority graphs over their individual/pairwise uses, indicating they all have important contributions to the superior performance of MOSGNN. 


\begin{figure}[h]
  \centering
\includegraphics[width=8.5cm]{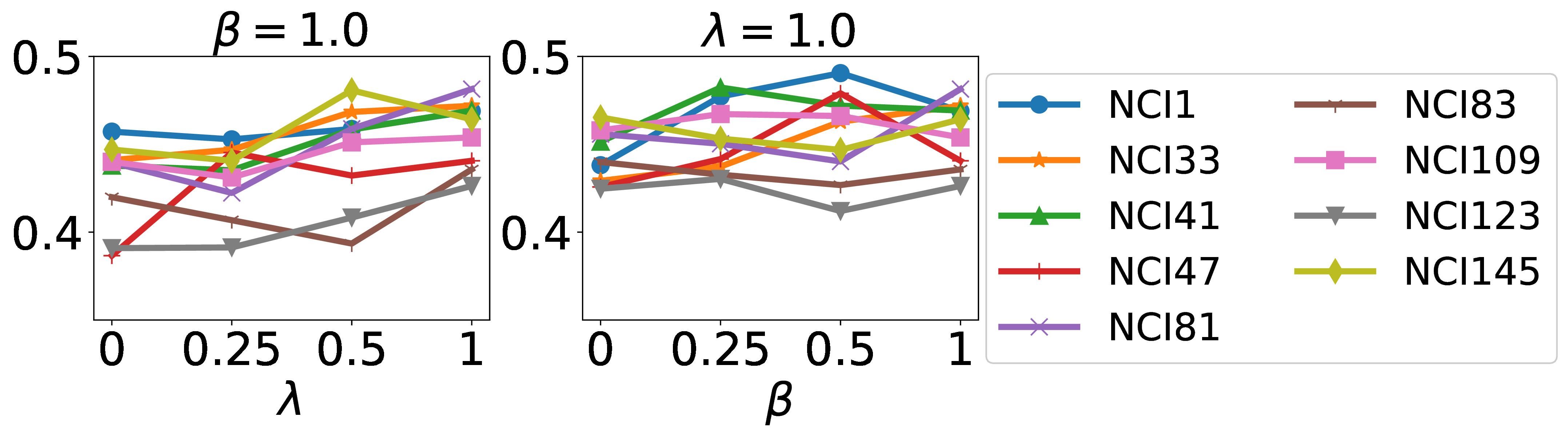}
  \caption{F1 scores (y-axis) of MOSGNN with different hyperparameter ($\lambda$ and $\beta$) settings on nine NCI datasets. }
  \label{hyperpara}
  \vspace{-0.2cm}
\end{figure} 

We also evaluate the effects of two hyperparameters ($\lambda$ and $\beta$) to the performance of MOSGNN. In Figure~\ref{hyperpara}, we show the results with $\beta=1.0$ fixed and varying $\lambda$ in $\{0, 0.25, 0.5, 1\}$, and vice versa. Clearly, larger $\lambda$ can bring better results in most datasets, which indicates that the information gained from the pairwise-graph-scale oversampling is general useful for the performance of MOSGNN. On the other hand, more efforts are required to tune the parameter $\beta$ for an effective use of the subgraph-scale oversampling. Nevertheless, both parameters can be well tuned on the validation dataset; the results here suggest that more careful tuning of $\beta$ is desired in practice.

\section{Conclusion}
This paper introduces a novel deep multi-scale oversampling approach MOSGNN for imbalanced graph classification. It can significantly extend the graph samples of the minority class with rich intra- and inter-graph semantics. These semantic-augmented minority graphs enable more effective training of GNNs on imbalanced graphs. 
The experiments show that MOSGNN learns significantly improved imbalanced graph classifiers over SOTA competing models 
and it can be also used by current imbalanced learning methods and GNN structures to significantly improve their classification performance.

\end{document}